\pdfoutput=1

\documentclass[11pt]{article}

\usepackage[]{emnlp2021}

\usepackage{times}
\usepackage{latexsym}

\usepackage[T1]{fontenc}

\usepackage[utf8]{inputenc}

\usepackage{microtype}
\usepackage{amsmath}
\usepackage{graphicx}

\title{Robustness Evaluation of Transformer-based Form Field Extractors \\ via Form Attacks}

\author{\textbf{Le Xue}~\thanks{\ \ Le and Mingfei contributed equally.} , \textbf{Mingfei Gao} $^*$, \textbf{Zeyuan Chen}, \textbf{Caiming Xiong and Ran Xu} \\
Salesforce Research, Palo Alto, USA \\
{\tt\small{\{lxue, mingfei.gao, zeyuan.chen, cxiong, ran.xu\}@salesforce.com}}
  }

\begin{document}
\maketitle
\begin{abstract}
We propose a novel framework to evaluate the robustness of transformer-based form field extraction methods via form attacks. We introduce 14 novel form transformations to evaluate the vulnerability of the state-of-the-art field extractors against form attacks from both OCR level and form level, including OCR location/order rearrangement, form background manipulation and form field-value augmentation. We conduct robustness evaluation using real invoices and receipts, and perform comprehensive research analysis. Experimental results suggest that the evaluated models are very susceptible to form perturbations such as the variation of field-values ($\sim$ 15\% drop in F1 score), the disarrangement of input text order($\sim$ 15\% drop in F1 score) and the disruption of the neighboring words of field-values($\sim$ 10\% drop in F1 score). Guided by the analysis, we make recommendations to improve the design of field extractors and the process of data collection.

\end{abstract}

\section{Introduction}
\begin{figure}
    \centering
    \includegraphics[width=1.0\linewidth]{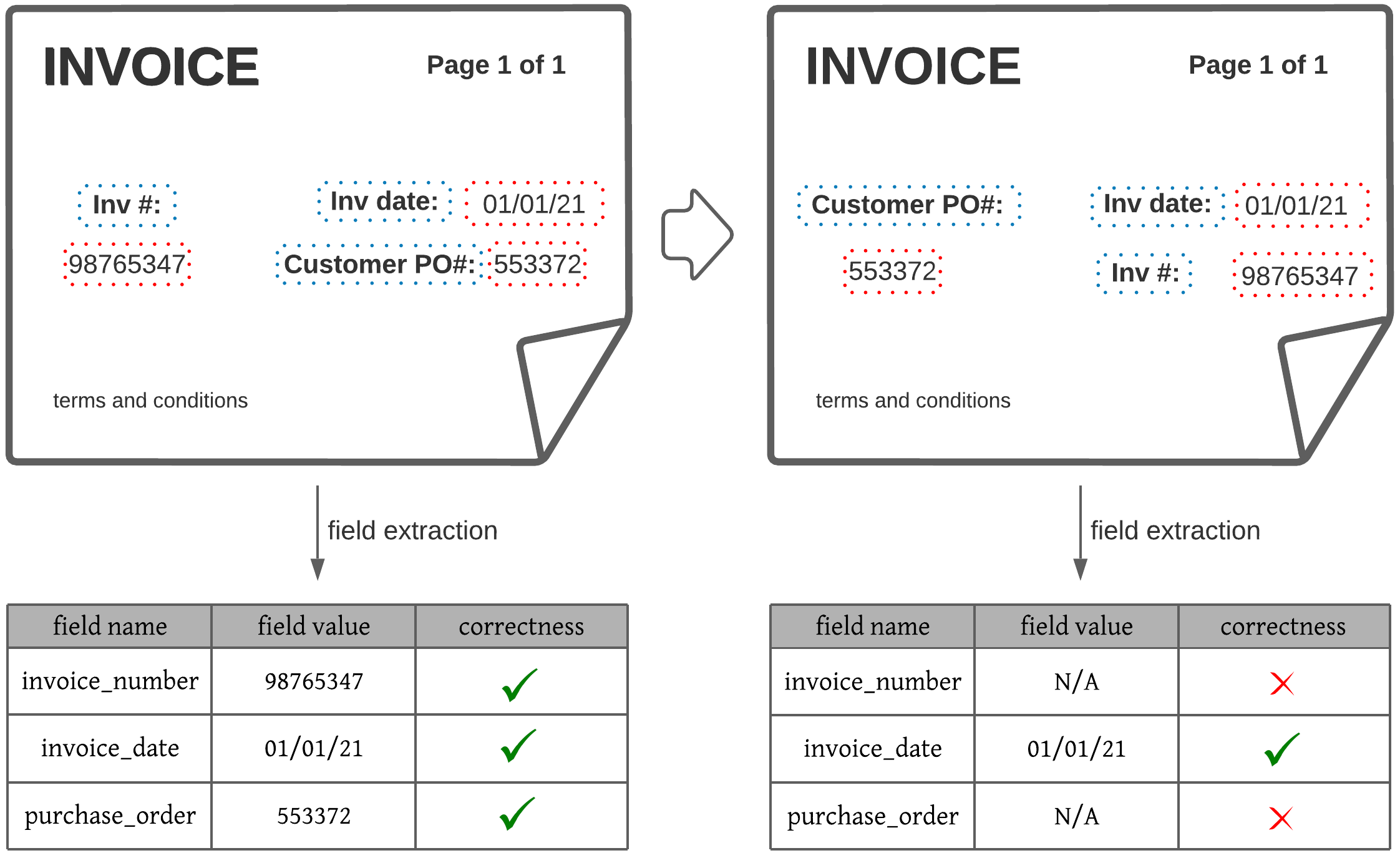}
    \caption{A form field extraction system may fail due to a slight modification to the form. Keys (concrete text-expressions of fields) are marked in blue boxes. Values are marked in red boxes.
    }
    \label{fig:idea}
\end{figure}

Forms such as invoices and receipts are essential in business workflows. Extracting target values for fields of interest from forms (see an example in Fig.~\ref{fig:idea}) is among the most important tasks in document understanding. There are large amounts of forms processed every day, but most current systems still rely on human labor to manually capture field-values from massively irrelevant information. Developing a method that automatically extracts field-values based on understanding the forms is crucial to reduce human labor, thus improve business efficiency.

Existing works~\cite{chiticariu-etal-2013-rule, 6628593, palm2019attend, majumder2020representation, xu2020layoutlm} focus on improving the modeling of field extractors and have made great progress. However, their evaluation paradigms are limited. First, most of the methods are evaluated using internal datasets. Internal datasets usually have very limited variations and are often biased towards certain data distributions due to the constraints of the data collection process. For example, the forms might be collected from just a few vendors in a relatively short time which leads to similar semantics and layouts across the forms. Second, public datasets lack for diversity in terms of both textual expression and form layouts. Take the most frequently used dataset, SROIE \cite{huang2019icdar2019}, as an example. The fields, \emph{company} and \emph{address}, are always on the very top in all receipts. Although the existing models achieve decent performance on these datasets, it is difficult to know whether they can generalize well. This issue can be solved by collecting large-scale diverse forms for evaluation, but it is very challenging since real forms usually contain customers' private information, thus are not publicly accessible. 

To tackle this dilemma, we propose a novel framework to evaluate the robustness of form field extractors by attacking the models using form transformations. We consider form perturbations from both OCR level and form level, including OCR text location/order rearrangement, form background manipulation, and form field-value augmentation. Fourteen form transformations are proposed to impose these attacks. Using the proposed framework, we conduct robustness analysis on two commonly used form types, i.e., invoices and receipts. Experimental results demonstrate that the state-of-the-art (SOTA) methods are particularly vulnerable to form perturbations, including the variation of field-values, the disarrangement of input text order, and the disruption of the neighboring words of field-values. Recommendations for model design and data collection/augmentation are made accordingly.

Our contributions are summarized as follows. First, we introduce a framework to measure the robustness of form field extractors by attacking the models using the proposed form transformations. To the best of our knowledge, this is the first work studying form attacks to field extraction methods. Second, we identify the susceptibilities of the SOTA methods by comprehensive robustness analysis on two form types using the proposed framework and make insightful recommendations.

\begin{figure*}
    \centering
    \includegraphics[width=1.0\linewidth]{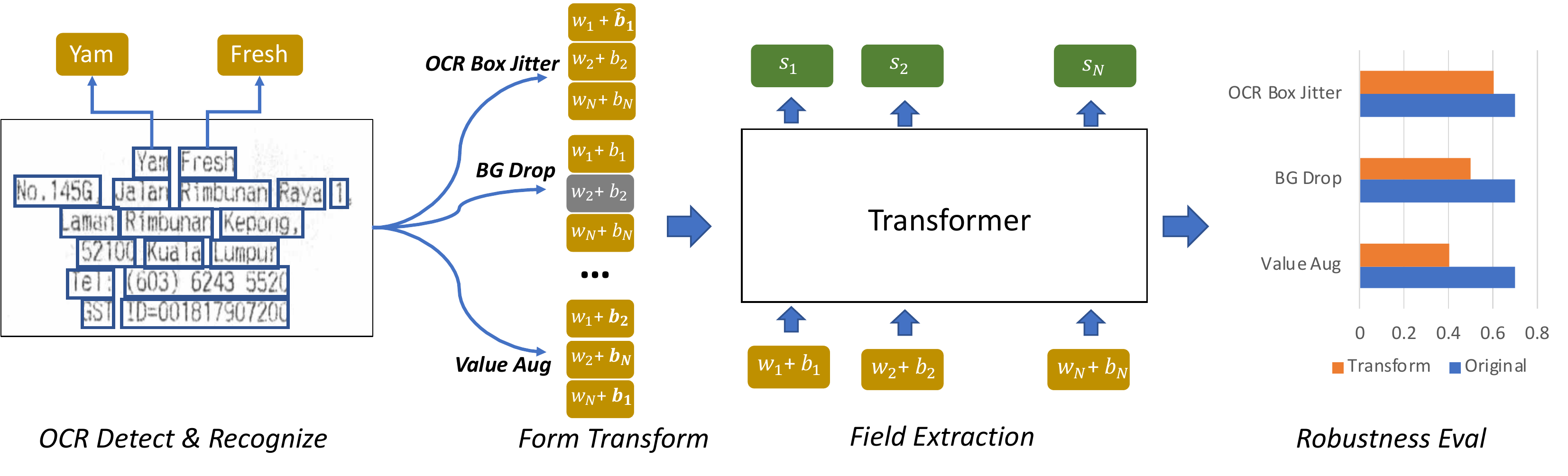}
    \caption{An illustration of our evaluation pipeline. An OCR engine is first used to extract texts and locations. 14 transformations are applied to the texts and their locations to generate diverse form variants. Then, each transformed set will input to a transformer-based field extractor. Robustness evaluation results are finally generated.}
    \label{fig:pipeline}
\end{figure*}
\section{Related Work}

\noindent\textbf{Information extraction from forms} is a widely researched area. \citet{katti2018chargrid} and \citet{denk2019bertgrid} encode each page of a form as a two-dimensional grid and extract header and line items from it using fully convolutional networks. DocStruct~\cite{wang2020docstruct} conducts document structure inference by encoding the form structure as a graph-like hierarchy of text fragments. Research works specifically focusing on form field extraction are more related to our work. Earlier methods~\cite{chiticariu-etal-2013-rule, 6628593} relied on pre-registered templates in the system for information extraction. \citet{palm2019attend} extract field-values of invoices via an Attend, Copy, Parse architecture. Recent methods formulate the field extraction problem as field-value pairing~\cite{majumder2020representation} and field tagging~\cite{xu2020layoutlm} tasks, where transformer~\cite{vaswani2017attention} based structures are used to extract informative form representation via modeling interactions among text tokens. We focus on evaluating transformer-based field extraction methods given their great predictive capability for the task.

\noindent\textbf{Robustness evaluation} of models has received considerable attention. Errudite~\cite{wu2019errudite} introduces model and task agnostic principles for informative error analysis of NLP models. \citet{ma2019nlpaug} propose NLPAug, which contains simple textual augmentations to improve model robustness. Some works aim at robustness of text attacks~\cite{morris2020textattack,zeng2020openattack,kiela2021dynabench}. A recent work, Robustness Gym~\cite{goel2021robustness}, presents a simple and extensible evaluation toolkit that unifies standard evaluation paradigms. There are also recent methods studying robustness of visual models~\cite{santurkar2020breeds,salman2020adversarially,taori2020measuring}. To the best of our knowledge, this work is the first one focusing on robustness evaluation of form field extraction systems.

\section{Preliminary: Transformer-based Form Field Extractor}
We are focusing on the robustness evaluation of transformer-based form field extractors due to their undisputedly outstanding performance. Before discussing the robustness evaluation, we first illustrate the field extraction pipeline.

In a standard field extraction system, an OCR engine is used to extract a set of words, $\{w_1, w_2, ..., w_N\}$ and their bounding box locations, $\{b_1, b_2,...,b_N\}$, where $N$ indicates the total number of words. Then, a transformer-based feature backbone is used to model the interactions between text tokens and generate informative token representations, $f_i$. Since both semantics and layouts are essential for field-value inference, we use LayoutLM~\cite{xu2020layoutlm} as the feature backbone. We also experiment with two more transformers, i.e., BERT~\cite{devlin2018bert} and RoBERTa~\cite{liu2019roberta} which only take text as input in the appendix. Finally, a fully connected (FC) layer is used to project the token features to field space and generate, $s_i=FC(f_i)$, where $s_i \in R^{1\times(M+1)}$ indicates the predicted field score and $M$ denotes the total number of positive fields. During training, cross entropy loss between $s_i$ and field label is utilized for model optimization. During inference, a post-processing method is applied to the predicted field scores to get the value for each field. We follow the simple criteria to generate field-values: (1) we find the predicted field label for each word by $\hat{fd}=\underset{c}{argmax}(s_{ic})$, where $c$ corresponds to fields (2) by default, for each field, we only keep the word as the value if its prediction score is the highest among all the words and larger than a threshold ($\theta=0.1$). For fields that often include multi-word values, e.g., address and company, we keep all the words exceeding the threshold and group nearby ones with the same predicted field. 

\noindent\textbf{Evaluation Metric}. 
End-to-end F1 score averaging over fields is used to evaluate models. We use exact string matching between the predicted values and the ground-truth to count true positives, false positives, and false negatives. Precision, recall and F1 scores are obtained accordingly for each field.

\section{Robustness Evaluation via Form Attacks}
We propose OCR level and form level transformations to attack field extractors. As shown in Fig.~\ref{fig:pipeline}, our transformations are performed after OCR extraction, and the transformed data is input to a transformer-based field extractor. An analysis is conducted via performance comparison between the original set and transformed sets. Each transformation and the principles behind it are introduced as follows.

\subsection{OCR Location and Order Rearrangement}
We transform the original data to meaningful variants by slightly altering the OCR locations and the text order arrangement. These transformations simulate scenarios where we may obtain different OCR results before inputting to field extractors due to various reasons, e.g., the quality of OCR engines.

\noindent\textbf{Center Shift and Box Stretch}. To evaluate model robustness to OCR text location jittering, we propose two word-level location transformations. In \emph{Center Shift}, we keep the box size and randomly shift the center of a box. The shifting is in proportional to the width (horizontally) and height (vertically) of the box, and the ratio is a random number drawn from a normal distribution, $\mathcal{N}(0,\delta_{cent})$. \emph{Box Stretch} randomly changes the four coordinates of a box in a similar way using $\mathcal{N}(0,\delta_{xy})$.

\noindent\textbf{Margin Padding}. Scanning forms may introduce white margins, which globally changes text locations. We use \emph{Margin Padding} to manipulate the locations of all the words in a form. We pad white margins in the left, right, up, and down sides of a form where the margin length is a generated random number between 1 and $r_{mp}$ of the page size.

\noindent\textbf{Global Shuffle}. We observe that organizing a transformer's inputs in reading order is particularly beneficial to understanding the form structure. However, the reading order is not always guaranteed by OCR engines. Hence, it is interesting to investigate model robustness to poor reading order quality. We use \emph{Global Shuffle} to shuffle the order of words before inputting to transformers. Note that the words and their locations are not changed at all and the only difference is the order of the word sequence input to the transformer.

\noindent\textbf{Neighbor Shuffle and Non-neighbor Shuffle}. Intuitively, local neighbors of a value make more contributions to its prediction. So, we propose \emph{Neighbor Shuffle} which shuffles the order of each value's neighbors and keeps the order of the rest. Oppositely, we also have \emph{Non-neighbor Shuffle}. A word, $w_i$, is defined as a value's neighbor if the IoU between $b_i$ and the neighbor zone of the value is larger than $0.5$. The neighbor zone is a box that shares the same center as the value box with expanded width and height (expand rate denoted as $r_{nb}$). We also include $n_{nb}$ nearby words from the original reading order as the neighbors.

\subsection{Form Background Manipulation}
 Form background generally affects the model performance in two ways: (1) some background words are strong indicators to improve field-values' recall and (2) accurate prediction of background reduces false positives, thus increase model precision. We propose the following transformations to evaluate model robustness to background perturbations. 
 
\noindent\textbf{BG Drop}. Background (BG) Drop mimics the scenario that some words are completely missed by OCR detection. This transformation removes background words together with the corresponding boxes at a probability of $p_{bgd}$.

\noindent\textbf{Neighbor BG Drop} is similar to \emph{Neighbor Shuffle}, which drops all background words if they are neighbors of a field-value.

\noindent\textbf{Key Drop}. Keys are concrete text-representations of fields in a form. For example, the field \emph{invoice\_number} may be represented as "INV \#", "Invoice No." etc. in a form. A key is a very important feature for value localization since the value is often located near the key. We propose \emph{Key Drop} to see the model performance change if keys are accidentally missed by OCR detection.

\noindent\textbf{BG Typo}. OCR recognition usually makes errors. \emph{BG Typo} simulates word-level string typos. We select each background word at a probability of $p_{typo}$. For each selected word, we apply one of the error types, including swapping, deleting, adding, and replacing a random or a specific character.~\footnote{We utilize the implementation of the string typos provided in https://pypi.org/project/typo/.}

\noindent\textbf{BG Synonyms}. Similar semantics may be represented using different word synonyms. \emph{BG Synonyms} randomly replaces each background word at a probability of $p_{bgs}$ with their synonyms.\footnote{We generate word synonyms using WordNet Interface (https://www.nltk.org/howto/wordnet.html).} 

\noindent\textbf{BG Adversarial}. Some forms contain only one word in the same data type as field-values. For example, there might be only one word with the date type of \emph{date}. It is less challenging for a model to recognize it as the \emph{invoice\_date}. However, this type of easy case is not always guaranteed in real-world applications. \emph{BG Adversarial} is used to increase the difficulty level by adding distraction. Concretely, we select background words at a probability of $p_{bga}$ and use adversarial words for replacement. For each replacement, we randomly choose a data type and then generate a random value of the corresponding data type. We focus on three data types, i.e., \emph{date}, \emph{number} and \emph{money}. We generate random dates using Faker.\footnote{ https://faker.readthedocs.io/en/master/} For numbers, we first generate the number length randomly and then a random number of the length accordingly. For \emph{money}, we obtain a random amount within lower and upper bounds. Then, we make the amount in money format, where we add a decimal point at the second to the left digit, place a comma at every third digit to the left of the decimal point and randomly insert $\$$ at the beginning. To protect strong indicators for values, neighbor words are not replaced.

\subsection{Form Field-Value Augmentations}
Modifying field-values is a more direct way to increase the diversity of the evaluation set. We augment field-values in both text and locations.

\noindent\textbf{Value Text Augment}. Field-values of forms may be biased due to the limitation of the data collection process. For example, the \emph{invoice\_date} may be restricted to the year the form is collected, and the \emph{invoice\_number} may be biased towards the vendor's numbering system. \emph{Value Text Augment} transformation targets at augmenting the field-values based on their data types. For each field-value, we randomly generate a substitute with the same data type following the same value generation procedure as we do in \emph{BG Adversarial}.

\noindent\textbf{Value Location Augment}. Form layouts can be very diverse in real-world scenarios. Intuitively, we should be able to infer a field-value as long as a key is represented properly no matter where we place the key-value pair in the document. We introduce~\emph{Value Location Augment} to increase layout diversity. To maintain the form format to the most, we keep the background as it is and shuffle the key-value pair's locations in the form. For example, for the field \emph{invoice\_number} (key: Invoice No., value: 1234) and \emph{invoice\_date} (key: Invoice date, value: 01/01/2021), we swap the box locations of ``Invoice No." and ``Invoice date", and also the locations of ``1234" and ``01/01/21".

\section{Experiments}
\label{sec:experiment}

We evaluate the robustness of transformation-based field extractors using our framework on two commonly used form types, i.e., invoices and receipts.

\subsection{Datasets}
Our evaluation models are trained using a labeled train set, and the best performing model is picked based on a validation set. We prepare a separate test set to perform the robustness evaluation. To perform the proposed transformations, we annotate both the key and value of each field of interest with their bounding box locations in a form.

\noindent\textbf{Invoice}. The train, valid, and test sets contain 158, 348 and 338 real invoices. They are collected from 111, 222, and 222 vendors, respectively. We sample at most 5 forms from the same vendor and the vendors of train, valid and test sets do not overlap. We consider 7 frequently used fields including \emph{invoice\_number}, \emph{purchase\_order}, \emph{invoice\_date}, \emph{due\_date}, \emph{amount\_due}, \emph{total\_amount} and \emph{total\_tax}. 

\noindent\textbf{Receipt}. We use the publicly available receipt dataset, SROIE. The annotations of their original test set are not publicly available, so we split the original train set to train, valid, and test sets based on their company names and sample at most 5 forms per company following~\citet{majumder2020representation}. Finally, we get 237 receipts for training, 76 for validation, and 74 for testing. The fields of interest are \emph{company}, \emph{address}, \emph{date} and \emph{total}. We add value boxes and annotate keys according to the text-level annotations provided by the original dataset.

\subsection{Implementation Details}
Our evaluation framework is implemented using Pytorch and the experiments are conducted on a single Tesla V100 GPU. The strength of a transformation is controlled by parameters. We set parameters to make moderate perturbations. In \emph{BG Typo}, \emph{BG Drop}, \emph{BG Synonyms} and \emph{BG Adversarial}, transformations are only applied to some selected words. We fix the pre-defined probabilities, $p_{typo}, p_{bgs}, p_{bga}$, to 0.1. $\theta_{center}$ is set to 0.5 and $\theta_{xy}$ is 0.1. $r_{mp}$ in Margin Padding is set to 0.3. When determining the value's neighbor zone, we set the expand rate as $r_{nb}=0.02$ and  $n_{nb}=2$.

We generate random values based on data types in \emph{BG Adversarial} and \emph{Value Text Augment}. For dates, we randomly pick a date from the year 2001 to 2021 in one of the formats, including mm/dd/yy, yy-mm-dd, dd/month/yy, and dd/mon/yy. For numbers, the number length is randomly generated from 3 to 12. The amount of money is randomly selected from 1 to 10,000,000.

We use a commercial OCR engine\footnote{https://api.einstein.ai/signup} for OCR extraction and utilize Tesseract\footnote{https://github.com/tesseract-ocr/tesseract} to rank the words in reading order. Our default transformer is LayoutLM~\cite{xu2020layoutlm} with text and boxes as inputs. We also evaluate using BERT~\cite{devlin2018bert} and RoBERTa~\cite{liu2019roberta} that take only text tokens as the inputs in Sec.~\ref{sec:appendix}. All models are finetuned from the corresponding base models. During training, we set the batch size to 8 and use the Adam optimizer with a learning rate of 5$e^{-5}$.

\subsection{Robustness Evaluation of Invoices}

\begin{table}[h]
\begin{tabular}{l|ccc}
\hline
 Transforms& Precision & Recall & F1 \\
\hline
Original & 70.9 & 69.3  & 70.0 \\
\hline
Center Shift & 70.8 & 69.9 & 70.3 \\
Box Stretch & 70.0 & 68.5 & 69.1 \\
Margin Pad & 69.6 & 68.3 & 68.9 \\
\hline
\end{tabular}
\caption{Evaluation of robustness to OCR location modifications on invoices. LayoutLM is used as the transformer model.}
\label{tab:invoice_layoutlm_ocr}
\end{table}

\noindent\textbf{OCR location modification}. Robustness evaluation of the LayoutLM model to OCR text location jittering is shown in Table~\ref{tab:invoice_layoutlm_ocr}. We obtain comparable results to the original performance when applying \emph{Center Shift} and \emph{Box Stretch} which indicates that slight box jittering is tolerable in our case. \emph{Margin Pad} shifts all text locations by adding random margins around a form. This transformation also just slightly decrease the model performance. 

\begin{table}[h]
\begin{tabular}{l|ccc}
\hline
 Transforms& Precision & Recall & F1 \\
\hline
Original & 70.9 & 69.3 & 70.0 \\
\hline
Global Shfl. & 58.5 & 53.9 & 55.9 \\
Neighbor Shfl. & 66.9 & 61.9 & 64.1 \\
Non-neighbor Shfl. & 67.0 & 65.3 & 65.9 \\
\hline
\end{tabular}
\caption{Evaluation of robustness to OCR text order on invoices. LayoutLM is used as the transformer model.}
\label{tab:invoice_layoutlm_shuffle}
\end{table}

\noindent\textbf{OCR text order} is essential to transformer-based field extractors since the order can serve as an important feature to improve model performance. We show the model performance when applying order shuffle to different places in Table~\ref{tab:invoice_layoutlm_shuffle}. The results show that if we shuffle all text orders, the performance drops dramatically by 14.1\% in F1 score. When only shuffle value neighbors, we obtain $\sim$6\% lower F1 score. We get $\sim$4\% lower F1 score if we shuffle non-neighbor words, even though we keep the text order of all values and their neighbors the same. The results demonstrate the importance of the text order. If the model is trained using texts with a good reading order, we may also want to ensure a good reading order during inference. 

A natural question is, what if we break the reading orders during training. Will this help us save the effort of ensuring test reading order during inference? We re-train field extractors using texts with random orders. We obtain 58.7\% in F1 score, which is 11.3\% lower than our baseline. The comparison result suggests that the text reading order is a very important feature. How to use it without overfitting to it is an interesting research topic.

\begin{table}[h]
\begin{tabular}{l|ccc}
\hline
 Transforms& Precision & Recall & F1 \\
\hline
Original & 70.9 & 69.3 & 70.0 \\
\hline
BG Drop & 69.8 & 66.2 & 67.7 \\
Neighbor BG Drop & 67.6 & 57.1 & 61.3 \\
Key Drop & 62.8 & 56.4 & 59.1\\
\hline
BG Typo & 69.6 & 66.9 & 68.1 \\
BG Synonyms & 70.4 & 68.8 &69.5  \\
BG Adversarial & 66.2 & 67.7 &66.9  \\
\hline
\end{tabular}
\caption{Evaluation of robustness to BG manipulations on invoices. LayoutLM is used as the transformer model.}
\label{tab:invoice_layoutlm_bg_drop}
\end{table}

\noindent\textbf{Background drop} related transformations simulate the scenarios where an OCR detector accidentally misses some background words. \emph{BG Drop} removes words randomly selected in the background. As shown in Table~\ref{tab:invoice_layoutlm_bg_drop}, global \emph{BG Drop} leads to slight performance decrease. When we apply \emph{Neighbor BG Drop}, the performance largely drops by $\sim$9\% in F1 score. Comparing \emph{BG Drop} and \emph{Neighbor BG Drop}, we find that neighbor words are indeed more important for value extraction. For a fair comparison, we have adjusted the dropping rate of \emph{BG Drop} to 0.13, such that the total number of dropped words is roughly equal to the number of neighbor words. Further, \emph{Key Drop} results in a similar performance decrease as \emph{Neighbor BG Drop}, although the total number of Keys is less than half of that for neighbor words.

\begin{table}[h]
\begin{tabular}{l|ccc}
\hline
 Transforms& Precision & Recall & F1 \\
\hline
Original & 70.9 & 69.3 & 70.0 \\
\hline
Value Text Aug. & 56.3 & 53.5 &54.5 \\
Value Location Aug. & 61.4 & 56.8 & 58.8 \\
\hline
\end{tabular}
\caption{Evaluation of robustness to background and field-value augmentations on invoices. LayoutLM is used as the transformer model.}
\label{tab:invoice_layoutlm_value_augment}
\end{table}

\noindent\textbf{Other background manipulations}. Some background words, e.g., keys and other useful indicators, are important features to localize values. Adding typos to the background words may be harmful to these good features, thus \emph{BG Typo} leads to a 2.4\% drop in recall rate.

Forms from different vendors may use different words even when represent similar semantics. We attack the models using \emph{BG Synonyms}. As shown in Table~\ref{tab:invoice_layoutlm_bg_drop}, our field extractors are quite robust to this transformation with only a negligible drop in F1 score.

\emph{BG Adversarial} is used to add background words (serve as distractions) with similar data types as field-values. As shown in Table~\ref{tab:invoice_layoutlm_bg_drop}, \emph{BG Adversarial} leads to 4.7\% drop in model precision.

\noindent\textbf{Value augmentations}. Field-values of forms may be limited in diversity. \emph{Value Text Augment} transformation augments field-values by replacing them with randomly generated values in the same data type. We augment the values of all the fields, except for \emph{total\_amount} and \emph{amount\_due}, since these two fields may involve complicated mathematical computations. For \emph{total\_tax}, we randomly select a number between 0 and 15\% of the \emph{total\_amount}. The comparison results in Table~\ref{tab:invoice_layoutlm_value_augment} show that the model performance drop significantly by 15.5\% F1 score.

\emph{Value Location Augment} changes the spatial arrangements of key-value pairs. In practice, we only shuffle the key-value pairs if they have the same number of key words and the same number of value words, resulting in more than 75\% key-value pairs relocated. The results in Table~\ref{tab:invoice_layoutlm_value_augment} demonstrate that \emph{Value Location Augment} significantly reduces F1 scores by 11.2\%.

\begin{figure}[h]
    \centering
    \includegraphics[width=1.0\linewidth]{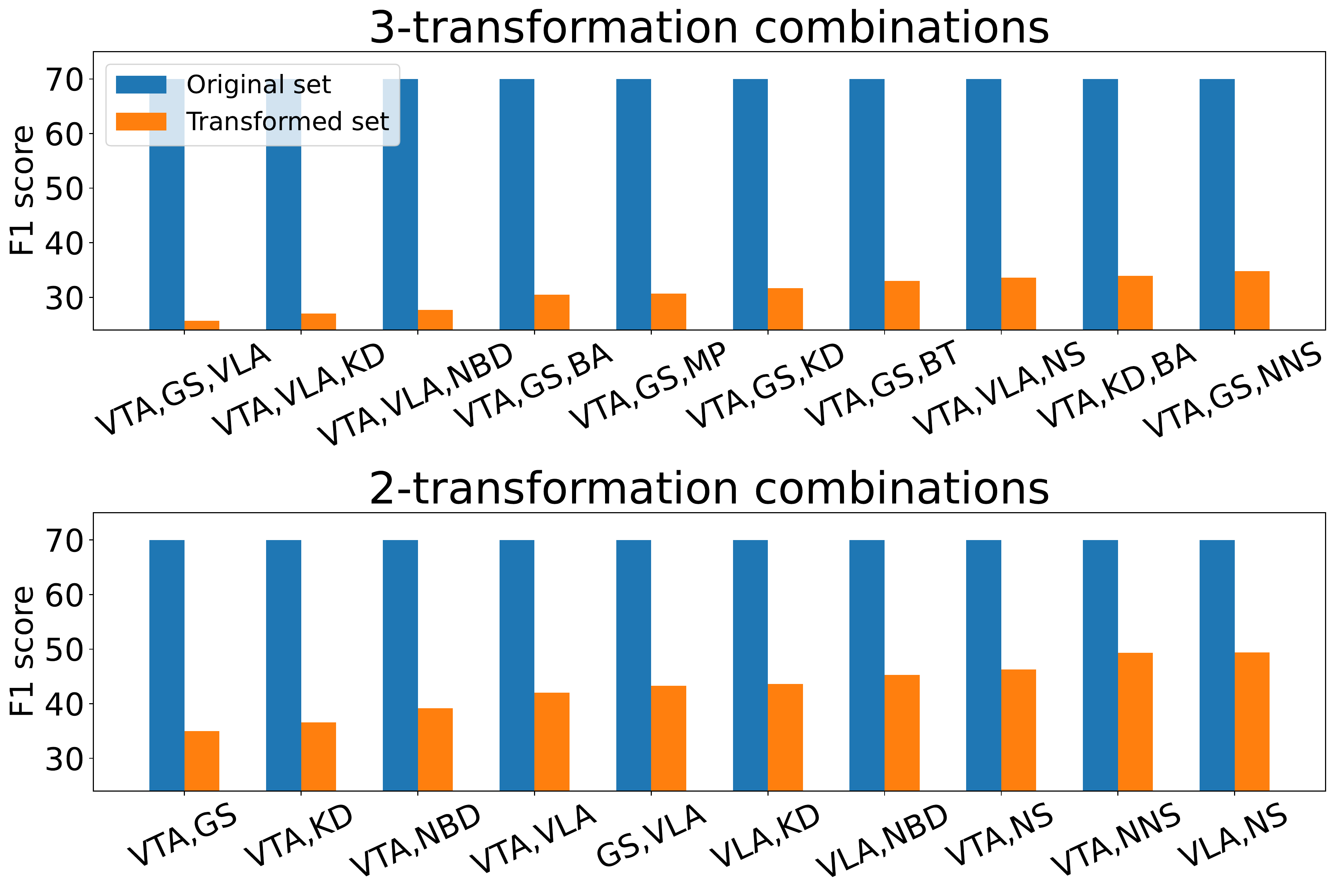}
    \caption{Top-10 most impactful 2-transformation and 3-transformation combinations. VTA: Value Text Augment, GS: Global Shift, VLA: Value Location Augment, KD: Key Drop, NBD: Neighbor Background Drop, BA: Background Adversarial, MP: Margin Pad, BT: Background Typo, NS: Neighbor Shuffle, NNS: Non-Neighbor Shuffle.}
    \label{fig:multiple transformations}
\end{figure}

\noindent\textbf{Multiple transformations}. The proposed transformations can be combined together to generate more diverse sets. We conduct exhaustive combinations of every two and three transformations which result in 91 2-transformation combinations and 364 3-transformation combinations.\footnote{We observe that different orders of transformations in a combination result in ignorable differences.} The top-10 most impactful combinations are shown in Fig.~\ref{fig:multiple transformations}. The comparison results suggest the following conclusions. 

\begin{figure*}
    \centering
    \includegraphics[width=1.0\linewidth]{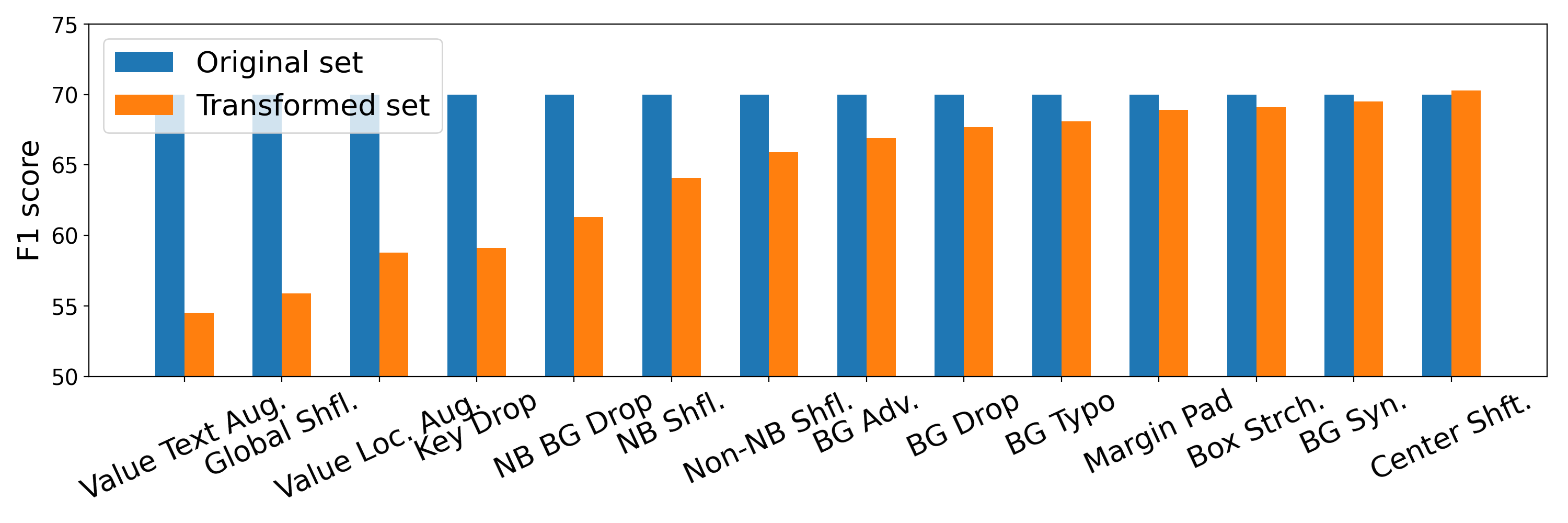}
    \caption{An overview of LayoutLM-based model performance drop due on different transformed dataset. The results are sorted by the performance gap.}
    \label{fig:comp_overview}
\end{figure*}

First, generally if an individual transformation drops more performance, it also contributes more drop when combined with other transformations. The most impactful combination is (\emph{Value text Augment}, \emph{Global Shuffle}, \emph{Value Location Augment}) with a F1 score of 25.7. They are the top-3 impactful transformations suggested in Fig.~\ref{fig:comp_overview}.

Second, some individual transformations are less impactful, but they affect more when combined with some specific transformations. For example, individual \emph{Margin Pad} ranks low in Fig.~\ref{fig:comp_overview}. However, it leads to more performance drop when combined with \emph{Value Text Augment} and \emph{Global Shuffle}. Their performance drop ranks 5 out of 364 combinations (F1 is 30.8).
This may due to that \emph{Margin Pad} (changes all words’ locations), \emph{Value Text Augment} (changes values’ texts) and \emph{Global Shuffle} (changes text input order) are three complementary transformations. When we do \emph{Margin Pad} alone, the model resorts to the information of value texts and text orders. However, when we do these three transformations together, the model becomes inevitably confused. 

Third, if the transformations have overlapping effects, their combination has a lower impact. For example, \emph{Key Drop}, \emph{Neighbor BG Drop} and \emph{Neighbor Shuffle} all manipulate neighbor words. The performance drop on their combination ranks 246 out of 364 combinations although their individual transformation is impactful (see Fig.\ref{fig:comp_overview}). The F1 score is 58.1 which is very close to an individual \emph{Key Drop} transformation (59.1).

\begin{figure}[h]
    \centering
    \includegraphics[width=1.0\linewidth]{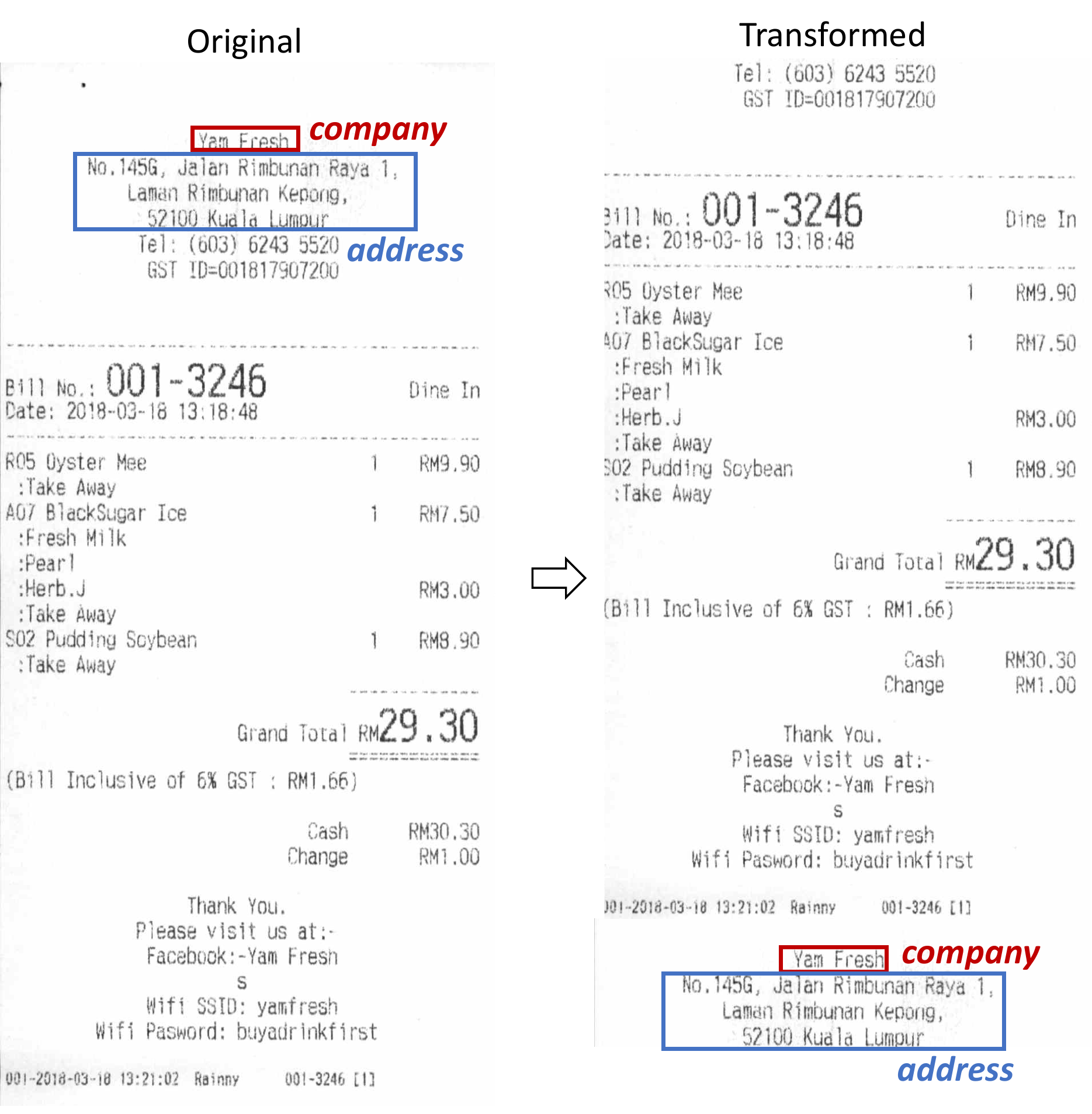}
    \caption{An illustration of \emph{Value Location Augment*} transformation on a receipt in SROIE dateset.}
    \label{fig:value_shift}
\end{figure}

\subsection{Robustness Evaluation of Receipts}
There are two interesting features of the SROIE dataset. First, a significant amount of field-values have no keys, for example, all values of \emph{company} and \emph{address}, and some values of \emph{date} and \emph{total}. Consequently, changing the context of values has a minor effect on model performance. Second, the layouts of different receipts are very fixed. For example, \emph{company} and \emph{address} are always on the very top of every receipt. So, models could easily overfit to field-value locations and the text order. As shown in Table~\ref{tab:receipt_layoutlm_augment}, \emph{Global Shuffle} leads to significant performance drop by 38.7\% F1 score. Specifically, the fields of \emph{address} and \emph{company} become 0\% F1 score when the text's order is completely shuffled before inputting to the transformer. The results demonstrate that the model is overfitting to the input text order, especially for \emph{address} and \emph{company}. 
\begin{table}[h]
\begin{tabular}{l|ccc}
\hline
 Transforms& Precision & Recall & F1 \\
\hline
Original & 81.8 & 80.1 & 80.9 \\
Global Shfl. & 42.5 & 41.9 &42.2 \\
Value Location Aug.* & 54.3 & 48.0 & 49.7\\
Value Text Aug. & 75.2 & 73.6 & 74.4\\
\hline
\end{tabular}
\caption{Robustness evaluation on SROIE dataset. LayoutLM is used as the transformer model.}
\label{tab:receipt_layoutlm_augment}
\end{table}

Most of the fields in SROIE have no keys. To augment receipt layouts, we design a dedicated method that locally moves field-value locations. Specifically, for \emph{company} and \emph{address} in SROIE receipts, we move the values to the bottom of the form and shift the rest above to fill the gap as shown in Fig.~\ref{fig:value_shift}. We refer to this transformation as \emph{Value Location Augment*}. This transformation changes the location of the values without breaking the text order within each value. We obtain 21.6\% and 1.7\% F1 score for \emph{company} and \emph{address}, respectively, which are around 60\% and 68\% lower than the original numbers.

Besides, we also evaluate models on test set transformed by \emph{Value Text Augment}. We replace values of \emph{company}, \emph{address} and \emph{date} using substitute randomly generated by Faker. Same as what we do for \emph{total\_amount} and \emph{amount\_due} for invoices, we keep the values of \emph{total} as they are. To maintain the layout structure, we only replace \emph{company} and \emph{address} if we are able to get a randomly generated sample with the same number of words as the original sample. This results in about 69\% \emph{company} values and 31\% \emph{address} values changed, respectively. As shown in Table~\ref{tab:receipt_layoutlm_augment}, the \emph{Value Text Augment} largely decreases the model performance by 5.5\% in F1 score.

\subsection{Observations and Suggestions}
An overview comparison of all the transformations of invoices is summarized in Fig.~\ref{fig:comp_overview}. As we can see, the top-3 substantial transformations are \emph{Value Text Augment}, \emph{Global Shuffle} and \emph{Value Location Augment}. Experiments on receipts also show the effectiveness of these three transformations.

We make the following recommendations based on the analysis. For data collection/augmentation, forms with more diverse values are preferable. For example, we may want \emph{dates} covering a wide range of time periods with more types of formats and \emph{numbers} being more extensive. Varying forms' layouts is also beneficial. Especially, we may want to focus on varying the arrangement of field-values instead of altering individual word locations locally. 

For the design of field extractors, we suggest making better utilization of the text order. As shown in our experiments, the text order is a very useful feature. How to utilize the text reading order without overfitting to it is an interesting topic. Besides, \emph{Key Drop} and \emph{Neighbor BG Drop} result in significant performance decreases as shown in Fig.~\ref{fig:comp_overview}. This suggests that value's neighbors, especially the keys, are essential for value extractions. Current state-of-the-art models use transformers to model interactions between all words. We believe paying attention to keys and neighbors in the model design has the potential to improve the existing field extraction systems.

\section{Conclusion and Future Work}
We proposed a novel framework to evaluate the robustness of transformer-based form field extractors via form attacks. We introduced 14 transformations that transform forms in different aspects, including OCR-level location and order, background contexts, and field-value text and layouts. We conducted studies on real invoices and receipts with three types of transformer-based models using our proposed framework. Research recommendations were made based on the robustness analysis. 

Improving field extraction from forms using the research analysis generated by the robustness evaluation is a very meaningful research area. The proposed transformations are potentially useful for increasing the diversity of training samples, thus improving model robustness. We will consider this in the future work.

\section{Broader Impact}
This work targets at robustness evaluation of form information extraction systems, so it has positive impacts such as identifying bias of existing information extractors and improving the fairness of model comparison. On the opposite side, our method may have unintended negative consequences in that we have proposed transformations that evaluate various aspects of model robustness, but the metrics we have selected may not be comprehensive. As a result,  there is likely some degree of model bias present that has been missed by the proposed framework. However, this negative impact is not specific to our work and should be considered in general in the field of robustness AI.

The invoice dataset is for internal use only and does not contain any personally identifiable data. The SROIE dataset is a public dataset under MIT license. All forms were annotated by the authors. Consequently, we are confident that the datasets do not have ethical issues.

\bibliography{anthology,custom}
\bibliographystyle{acl_natbib}
\clearpage

\appendix

\section{Appendix}
\label{sec:appendix}

\renewcommand{\thefigure}{A\arabic{figure}}
\setcounter{figure}{0}

\setcounter{table}{0}
\renewcommand{\thetable}{A\arabic{table}}
\subsection{More Robustness Evaluation of Invoices}
The results of BERT and RoBERTa on invoices are summarized in Table~\ref{tab:invoice_bert_augment} and Table~\ref{tab:invoice_roberta_augment}.

\begin{table}[h]
\begin{tabular}{l|ccc}
\hline
 Transforms& Precision & Recall & F1 \\
\hline
Original & 58.4 & 58.1 & 57.8\\
\hline
Global Shfl. & 40.5 & 38.2 & 38.6\\
Neighbor Shfl. & 51.5 & 50.7 & 50.6 \\
Non-neighbor Shfl. & 57.9 & 56.4 & 56.6\\
\hline
BG Drop & 57.7 & 57.2 & 56.9\\
Neighbor BG Drop & 49.9 & 44.7 & 46.7\\
Key Drop & 50.2 & 46.5 & 47.6\\
\hline
BG Typo & 56.0 & 54.1 & 54.7 \\
BG Synonyms & 58.8 & 58.3 & 58.0\\
BG Adversarial & 48.7 & 51.6 & 49.7\\
\hline
Value Text Aug. & 36.2 & 33.8 & 34.3\\
Value Location Aug. & 55.5 & 54.4 & 54.4\\
\hline
\end{tabular}
\caption{Robustness evaluation on invoice dataset. BERT is used as the transformer model.}
\label{tab:invoice_bert_augment}
\end{table}

\begin{table}[h]
\begin{tabular}{l|ccc}
\hline
 Transforms& Precision & Recall & F1 \\
\hline
Original & 63.4 & 59.1 & 61.1 \\
\hline
Global Shfl. & 40.5 & 33.3 & 36.4\\
Neighbor Shfl. & 58.9 & 53.2 & 55.9 \\
Non-neighbor Shfl. & 60.6 & 57.0 & 58.7\\
\hline
BG Drop & 63.6 & 58.8 & 61.0\\
Neighbor BG Drop & 58.8 & 49.5 & 53.5\\
Key Drop & 57.7 & 49.8 & 53.2\\
\hline
BG Typo & 62.1 & 57.7 & 59.8 \\
BG Synonyms & 62.7 & 58.5 & 60.4\\
BG Adversarial & 56.2 & 55.2 & 55.5\\
\hline
Value Text Aug. & 44.9 & 39.2 & 41.5\\
Value Location Aug. & 57.9 & 53.9 & 55.7\\
\hline
\end{tabular}
\caption{Robustness evaluation on invoice dataset. RoBERTa is used as the transformer model.}
\label{tab:invoice_roberta_augment}
\end{table}

\noindent\textbf{OCR text order}. BERT and RoBERTa do not have text locations as inputs, so they rely more on text order than LayoutLM does. When deploying \emph{Global Shuffle}, we observe more performance drop, i.e., by 19.2\% F1 score for BERT and by 24.7\% F1 score for RoBERTa.

\noindent\textbf{Background drop}. We observe similar results on BERT and RoBERTa as compared to that on LayoutLM. \emph{Key Drop} and \emph{Neighbor BG Drop} have more impact than global \emph{BG Drop}.

\noindent\textbf{Other background manipulations}. We observe 3.1\% and 1.3\% drop in F1 score on \emph{BG Typo}, when using BERT and RoBERTa. Similar to LayoutLM, BERT and RoBERTa are robust to \emph{BG Synonyms}. \emph{BG Adversarial} leads to 9.7\%  and 7.2\% drop in model precision for BERT and RoBERTa based methods, respectively.

\noindent\textbf{Value augmentations}. \emph{Value Text Augment} results in 23.5\% (BERT) and 19.6\% (RoBERTa) drop in F1 score. Since BERT and RoBERTa do not rely on location input, the performance drop on \emph{Value Location Augment} of these two models is much less than that of the LayoutLM (drop 11.2\%). 

\subsection{More Robustness Evaluation of Receipts}
The results of BERT and RoBERTa on receipts are summarized in Table~\ref{tab:receipt_bert_augment} and Table~\ref{tab:receipt_roberta_augment}. The experimental results suggest that these three transformations have significant impact to the performance of BERT and RoBERTa.

\begin{table}[h]
\begin{tabular}{l|ccc}
\hline
 Transforms& Precision & Recall & F1 \\
\hline
Original & 75.4 & 73.3 & 74.3 \\
Global Shfl. & 37.9 & 36.1 & 37.0 \\
Value Location Aug.* & 51.9 & 44.9 & 46.0\\
Value Text Aug. & 71.2 & 69.2 & 70.2\\
\hline
\end{tabular}
\caption{Robustness evaluation on SROIE dataset. BERT is used as the transformer model.}
\label{tab:receipt_bert_augment}
\end{table}

\begin{table}[h]
\begin{tabular}{l|ccc}
\hline
 Transforms& Precision & Recall & F1 \\
\hline
Original & 77.3 & 74.7 & 75.9 \\
Global Shfl. & 40.6 & 37.1 & 38.8 \\
Value Location Aug.* & 48.6 & 41.9 & 43.4\\
Value Text Aug. & 72.5 & 70.6 & 71.5\\
\hline
\end{tabular}
\caption{Robustness evaluation on SROIE dataset. RoBERTa is used as the transformer model.}
\label{tab:receipt_roberta_augment}
\end{table}
\end{document}